\PassOptionsToPackage{dvipsnames}{xcolor}
\documentclass[9pt,conference]{IEEEtran}

\ifCLASSOPTIONcompsoc
    \usepackage[caption=false, font=normalsize, labelfont=sf, textfont=sf]{subfig}
\else
\usepackage[caption=false, font=footnotesize]{subfig}
\fi
\usepackage{graphicx}
\usepackage{booktabs}
\usepackage{url}
\usepackage{multirow}
\usepackage[countmax]{subfloat}
\usepackage{comment}
\usepackage{orcidlink}
\usepackage{tikz}
\usepackage{gensymb}
 \usepackage[numbers]{natbib}

\usepackage{amsmath}
\usepackage{amsfonts}
\usepackage{amssymb}
\usepackage{algorithm}

\usepackage{algpseudocode}

\usepackage[font=small,skip=5pt]{caption}

\usepackage{caption}

\begin{document}

\title{Secure and Storage-Efficient Deep Learning Models for Edge AI Using Automatic Weight Generation}

\author{
    \IEEEauthorblockN{Habibur Rahaman, Atri Chatterjee, and Swarup Bhunia}  
    \IEEEauthorblockA{School of Electrical and Computer Engineering, University of Florida \\  
    \{rahaman.habibur, a.chatterjee\}@ufl.edu, swarup@ece.ufl.edu}  
}

\maketitle

\begin{abstract}
Complex neural networks require substantial memory to store a large number of synaptic weights. This work introduces \textit{WINGs} (Automatic Weight Generator for Secure and Storage-Efficient Deep Learning Models), a novel framework that dynamically generates layer weights in a fully connected neural network (FC) and compresses the weights in convolutional neural networks (CNNs) during inference, significantly reducing memory requirements without sacrificing accuracy. WINGs framework uses principal component analysis (PCA) for dimensionality reduction and lightweight support vector regression (SVR) models to predict layer weights in the FC networks, removing the need for storing full-weight matrices and achieving substantial memory savings. It also preferentially compresses the weights in low-sensitivity layers of CNNs using PCA and SVR with sensitivity analysis. The sensitivity-aware design also offers an added level of security, as any bit-flip attack with weights in compressed layers has an amplified and readily detectable effect on accuracy. WINGs achieves 53x compression for the FC layers and 28x for AlexNet with MNIST dataset, and 18x for Alexnet with CIFAR-10 dataset 
with 1-2\% accuracy loss. This significant reduction in memory results in higher throughput and lower energy for DNN inference, making it attractive for resource-constrained edge applications.



\end{abstract}


\renewcommand\IEEEkeywordsname{Keywords}

%

\vspace{-3pt}
\IEEEpeerreviewmaketitle
\vspace{-5pt}
\section{Introduction}
The success of large artificial intelligence (AI) and machine learning models with millions or billions of parameters (weights) comes at the cost of high memory demands and computational complexity. Implementing large DNN models on resources-constrained systems, such as mobiles, Internet of Things (IoT) end-point/edge devices, or smart sensors poses significant challenges due to high memory and computational demands \cite{han2016eie}.

Most deep learning models rely on static storage of weight matrices, which are loaded into memory during inference. As model sizes grow, this approach leads to an unsustainable demand for memory, especially for Deep neural network (DNN) architectures, like BERT \cite{devlin2018bert}, GPT \cite{brown2020language}, or ResNet \cite{he2016deep}. Table \ref{tab:my-table-1} shows the memory requirement and memory access energy for large DNN architectures and indicates that they contain millions/billion parameters, requiring hundreds of gigabytes of memory to store model weights alone, making it infeasible for deployment on devices with limited memory capacity. On-chip SRAM alone cannot store all weights, necessitating external memory to store complete weight matrices. The latency and increased memory access energy of the off-chip memory are bottlenecks to achieving the performance of large DNN. Thus, reducing the memory footprint of DNN is essential for efficient deployment in real-life embedded applications.

Existing solutions, which aim at directly or indirectly reducing memory requirements, such as \textit{Quantization}, \cite{rastegari2016xnor}, \textit{Pruning} \cite{zhu2017prune}, \textit{low-rank factorization} \cite{sainath2013low}, \textit{hashing} methods \cite{chen2015compressing}, MobileNets \cite{howard2017mobilenets}, SqueezeNet \cite{iandola2016squeezenet}, etc., can reduce storage needs, often at the cost of accuracy, requiring retraining, or incurring large hardware overhead for decoding/decompression. \textit{On-the-fly weight  generation} techniques \cite{ha2017hypernetworks} reduce memory usage by dynamically generating weights during inference. Most of the existing memory reduction techniques, however, require retraining to maintain accuracy.



\begin{table} [h]
\centering
\caption{Memory and energy requirements for large DNN architectures (access energy for DDR3 and on-chip SRAM are 70pJ/bit and 0.16pJ/bit, respectively [28nm CMOS technology] \cite{ko2017adaptive}}
\label{tab:my-table-1}
\resizebox{\columnwidth}{!}{%
\begin{tabular}{|l|c|cc|cccl|}
\hline
\multicolumn{1}{|c|}{\multirow{3}{*}{\textbf{\begin{tabular}[c]{@{}c@{}}Model \\ (Architecture)\end{tabular}}}} &
  \multirow{3}{*}{\textbf{Parameters}} &
  \multicolumn{2}{c|}{\textbf{\begin{tabular}[c]{@{}c@{}}Weight Memory \\ Requirement\end{tabular}}} &
  \multicolumn{4}{c|}{\textbf{Memory Access Energy (Joules)}} \\ \cline{3-8} 
\multicolumn{1}{|c|}{} &
   &
  \multicolumn{1}{c|}{\multirow{2}{*}{\textbf{32-bit}}} &
  \multirow{2}{*}{\textbf{16-bit}} &
  \multicolumn{2}{c|}{\textbf{\begin{tabular}[c]{@{}c@{}}32-bit weight \\ access\end{tabular}}} &
  \multicolumn{2}{c|}{\textbf{\begin{tabular}[c]{@{}c@{}}16-bit weight \\ access\end{tabular}}} \\ \cline{5-8} 
\multicolumn{1}{|c|}{} &
   &
  \multicolumn{1}{c|}{} &
   &
  \multicolumn{1}{c|}{\textbf{DDR3}} &
  \multicolumn{1}{c|}{\textbf{SRAM}} &
  \multicolumn{1}{c|}{\textbf{DDR3}} &
  \multicolumn{1}{c|}{\textbf{SRAM}} \\ \hline
Bert (110MB) & 110 million & \multicolumn{1}{c|}{440 MB}  & 220 MB & \multicolumn{1}{c|}{0.2584}    & \multicolumn{1}{c|}{0.0006} & \multicolumn{1}{c|}{0.1292}   & 0.0003 \\ \hline
Bert (340MB) & 340 million & \multicolumn{1}{c|}{1.32 GB} & 680 MB & \multicolumn{1}{c|}{0.8178}    & \multicolumn{1}{c|}{0.0019} & \multicolumn{1}{c|}{0.3993}   & 0.0009 \\ \hline
GPT-3(13B)   & 13 Billion  & \multicolumn{1}{c|}{52 GB}   & 26 GB  & \multicolumn{1}{c|}{31.2674}   & \multicolumn{1}{c|}{0.0715} & \multicolumn{1}{c|}{15.6337}  & 0.0357 \\ \hline
GPT-3(175B)  & 175 Billion & \multicolumn{1}{c|}{700GB}   & 350 GB & \multicolumn{1}{c|}{420.9068}  & \multicolumn{1}{c|}{0.9621} & \multicolumn{1}{c|}{210.4534} & 0.4810 \\ \hline
GPT-4(200B)  & 200 Billion & \multicolumn{1}{c|}{800 GB}  & 400 GB & \multicolumn{1}{c|}{481.0363}  & \multicolumn{1}{c|}{1.0995} & \multicolumn{1}{c|}{210.5182} & 0.5498 \\ \hline
GPT-4(500B)  & 500 Billion & \multicolumn{1}{c|}{2 TB}    & 1 TB   & \multicolumn{1}{c|}{1231.4530} & \multicolumn{1}{c|}{2.8147} & \multicolumn{1}{c|}{615.7265} & 1.4074 \\ \hline
\end{tabular}%
}
\end{table}




\begin{figure}[h]
    \centering
    \includegraphics[width=\linewidth]{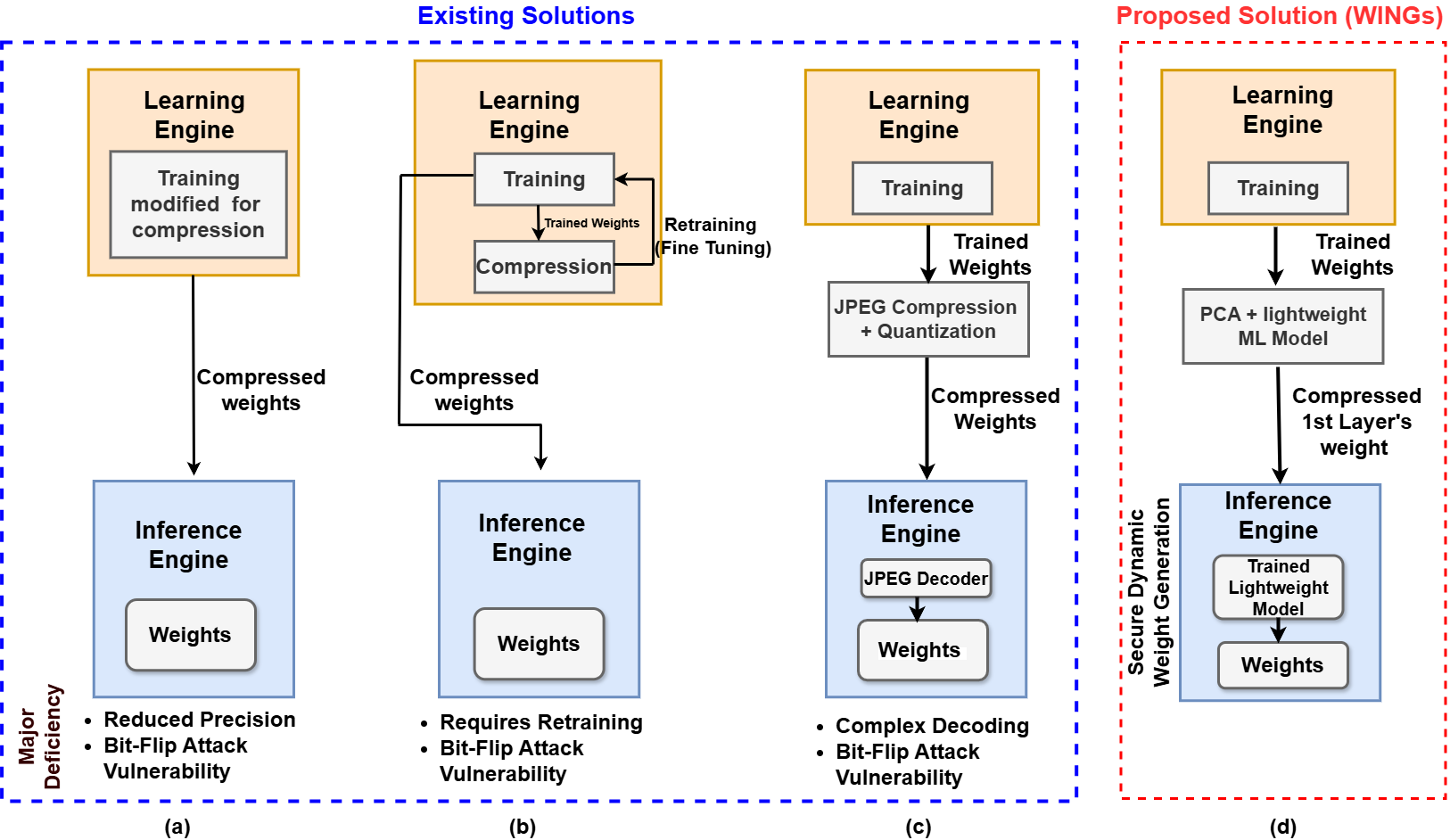}
    \caption{This figure describes the overall weight generation/compression schemes for a scalable DNN. (a) This shows a compression mechanism that requires modified training algorithm \cite{courbariaux2016binarized}, (b) This shows compression mechanism that requires fine tuning \cite{han2015deep}, (c) This shows non-scalabile JPEG based compression scheme \cite{ko2017adaptive}, (d) This shows weight generation and security aware (WINGs) scheme.}
    \label{fig:Comparision of Schemes}
\end{figure}

\begin{table}[]
\centering
\caption{Comparison of Weight Compression Techniques}
\label{tab:my-table-11}
\resizebox{\columnwidth}{!}{%
\begin{tabular}{|c|c|c|c|c|}
\hline
\textbf{Attribute} &
  \textbf{\begin{tabular}[c]{@{}c@{}}Quantized Neural \\ Networks (QNN) \cite{courbariaux2016binarized}\end{tabular}} &
  \textbf{\begin{tabular}[c]{@{}c@{}}Deep \\ Compression (DC)\cite{han2015deep} \end{tabular}} &
  \textbf{\begin{tabular}[c]{@{}c@{}}JPEG \\ Compression\cite{ko2017adaptive}\end{tabular}} &
  \textbf{WINGs (Proposed)} \\ \hline
\textbf{\begin{tabular}[c]{@{}c@{}}Energy \\ Overhead\end{tabular}} &
  \begin{tabular}[c]{@{}c@{}}Low; replaces arithmetic \\ operations with bitwise (XNOR, \\ popcount) operations \\ for high efficiency.\end{tabular} &
  \begin{tabular}[c]{@{}c@{}}Moderate; \\ retraining for sparse \\ weights adds overhead.\end{tabular} &
  \begin{tabular}[c]{@{}c@{}}High; significant energy \\ due to inverse DCT \\ and quantization reversal.\end{tabular} &
  \begin{tabular}[c]{@{}c@{}}Minimal; PCA + SVR \\ are computationally \\ lightweight and ideal \\ for IoT and mobile devices.\end{tabular} \\ \hline
\textbf{Mechanism} &
  \begin{tabular}[c]{@{}c@{}}Dynamic quantization of\\  weights/activations \\ (binary/low-precision\\  values) during\\  training and inference.\end{tabular} &
  \begin{tabular}[c]{@{}c@{}}Pruning, quantization,\\  and Huffman coding \\ for static weight compression.\end{tabular} &
  \begin{tabular}[c]{@{}c@{}}JPEG DCT encoding \\ with adaptive quality \\ factors for static \\ compression.\end{tabular} &
  \begin{tabular}[c]{@{}c@{}}PCA for dimensionality \\ reduction and \\ SVR for dynamic, \\ on-the-fly weight generation.\end{tabular} \\ \hline
\textbf{\begin{tabular}[c]{@{}c@{}}Real-Time \\ Use\end{tabular}} &
  \begin{tabular}[c]{@{}c@{}}Suitable; low-latency \\ execution with hardware \\ optimizations \\ (e.g., XNOR kernels).\end{tabular} &
  \begin{tabular}[c]{@{}c@{}}Limited; requires \\ retraining and \\ preprocessing.\end{tabular} &
  \begin{tabular}[c]{@{}c@{}}Pre-decoding weights \\ introduces latency, \\ unsuitable for \\ real-time applications.\end{tabular} &
  \begin{tabular}[c]{@{}c@{}}Highly suitable; \\ dynamic weight generation \\ ensures low-latency and \\ real-time performance.\end{tabular} \\ \hline
\textbf{\begin{tabular}[c]{@{}c@{}}Accuracy \\ Handling\end{tabular}} &
  \begin{tabular}[c]{@{}c@{}}Slight degradation (e.g., 1–2\% \\ on ImageNet) due \\ to binarization/\\ low-precision quantization.\end{tabular} &
  \begin{tabular}[c]{@{}c@{}}Retraining ensures \\ no accuracy loss \\ for compressed networks.\end{tabular} &
  \begin{tabular}[c]{@{}c@{}}Entropy-based \\ compression risks \\ higher accuracy \\ degradation.\end{tabular} &
  \begin{tabular}[c]{@{}c@{}}Selective compression \\ of low-sensitivity \\ layers ensures \\ minimal accuracy\\  loss (1–2\%).\end{tabular} \\ \hline
\textbf{Security} &
  \begin{tabular}[c]{@{}c@{}}No explicit security; \\ benefits from \\ smaller parameter\\  space.\end{tabular} &
  \begin{tabular}[c]{@{}c@{}}No explicit \\ security features.\end{tabular} &
  \begin{tabular}[c]{@{}c@{}}Vulnerable to \\ bit-flip attacks;\\  lacks security \\ mechanisms.\end{tabular} &
  \begin{tabular}[c]{@{}c@{}}Sensitivity-aware design \\ amplifies \\ tamper errors, enabling \\ robust attack detection.\end{tabular} \\ \hline
\textbf{\begin{tabular}[c]{@{}c@{}}Compression \\ Ratio\end{tabular}} &
  \begin{tabular}[c]{@{}c@{}}Up to 32x memory \\ reduction \\ with minimal \\ accuracy loss.\end{tabular} &
  \begin{tabular}[c]{@{}c@{}}35–49x \\ (AlexNet/VGG-16), \\ no accuracy loss.\end{tabular} &
  \begin{tabular}[c]{@{}c@{}}Moderate; 63.4x (MLPs) \\ and 31.3x (CNNs),\\  but higher accuracy loss.\end{tabular} &
  \begin{tabular}[c]{@{}c@{}}53x (FCNs) and \\ 29x (CNNs) with \\ only 1–2\% \\ accuracy loss.\end{tabular} \\ \hline
\end{tabular}%
}
\end{table}


Further, AI hardware used in edge applications are vulnerable to diverse adversarial attacks on the trained model. Among them, bit-flip attacks have emerged as a security concern for intelligent computing at the edge, due to physical access of edge devices. Although existing solutions, such as PARA \cite{kim2014flipping}, ECC \cite{cojocar2019exploiting}, and DNN-based protection \cite{li2020defending}, aim at addressing this vulnerability, they generally incur high overhead and unacceptable protection. 


 PARA  \cite{kim2014flipping},  requires approximately 20\% timing overhead (memory access latency) and additional logic 5\%-15\%. Similarly, ECC \cite{cojocar2019exploiting} protection needs around 12. 5\% compaction overhead, 10\% timing overhead, and around 20\% hardware overhead.

\begin{figure*}[!h]
    \centering
    \includegraphics[width=1\textwidth]{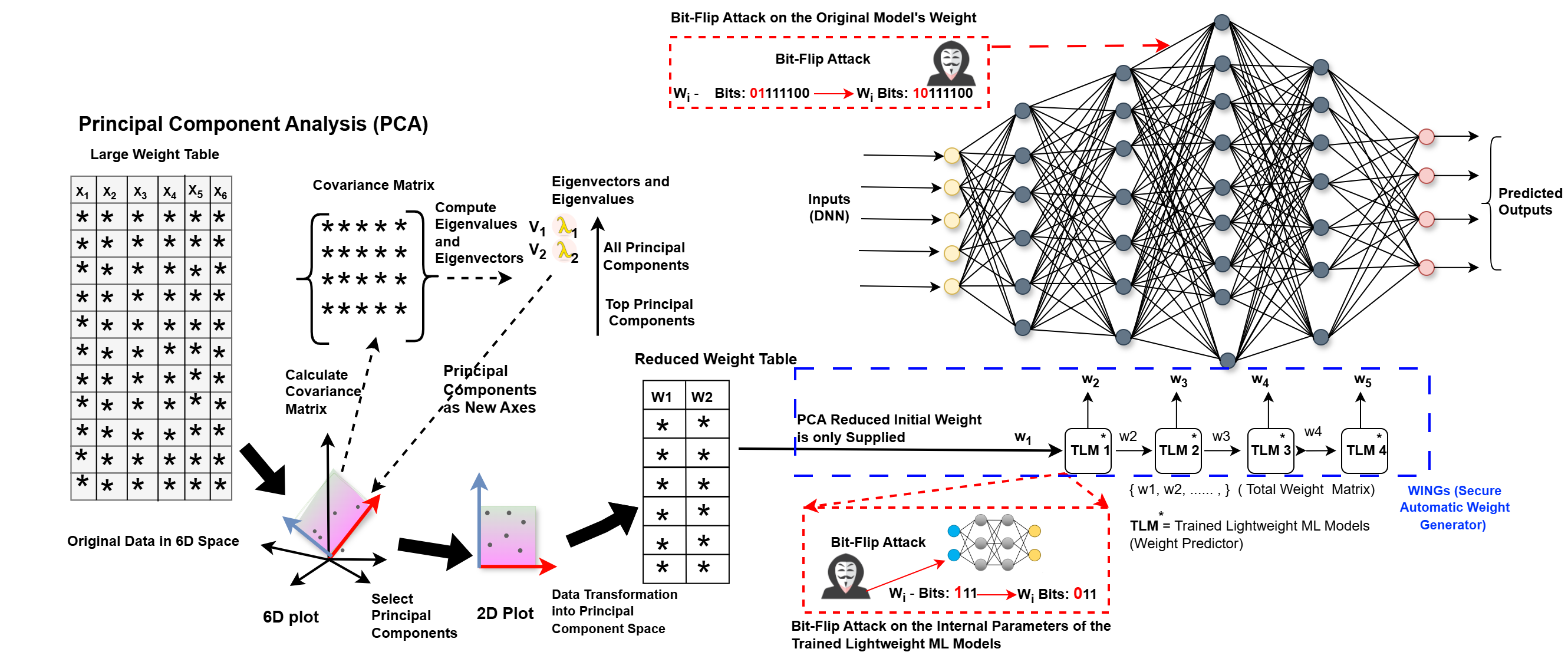}
    \caption {Flow diagram illustrating automatic weight generation and compression in the WINGs framework: This figure demonstrates Principal Component Analysis (PCA) for dimensionality reduction, support vector regression (SVR) for lightweight weight prediction, and sensitivity-aware compression for enhanced memory efficiency and security in Deep Neural Networks.}
    \label{fig:Lightweight}
\end{figure*}

In this paper, we introduce \textit{WINGs} that simultaneously address the weight storage efficiency and resistance to bit-flip attacks for DNN models. Figure \ref{fig:Comparision of Schemes} and table \ref{tab:my-table-11} illustrate the weight construction procedure used alongside the proposed idea. \textit{WINGs} is compared with alternative weight generation and compression schemes, as shown in Figure \ref{fig:Comparision of Schemes} and table \ref{tab:my-table-11}. \textit{WINGs} dynamically generates layer weights during inference and aims to significantly minimize memory usage without sacrificing accuracy. The key innovation of this work lies in the use of  \textit{PCA} \cite{pearson1901pca} to reduce the dimensionality of the weight matrices and the use of the lightweight ML model SVR to predict the weights for subsequent layers. Instead of storing the entire weight matrix, it retains a compact representation of the weights, as illustrated in figure \ref{fig:Lightweight}, allowing dynamic generation of the required weights for each layer during inference. The model generates the weights needed only at each specific moment.

\vspace{-5pt}

 \subsection{Key Contributions}
The main contributions of this work are summarized as follows:
\begin{itemize}
    \item We propose a framework, \textbf{WINGs} that compresses and dynamically generates layer weights during inference, eliminating the need to store full-weight matrices, resulting in a significant reduction in memory usage.
    \item The framework leverages \textbf{PCA} for dimensionality reduction and uses \textbf{(SVR)} to predict and reconstruct compressed weights efficiently.
    \item We introduce a \textbf{sensitivity-aware compression} technique that uses gradient-based sensitivity analysis to selectively compress low-sensitivity layers while preserving high-sensitivity layers in their original form to maintain model accuracy.
    \item This also improves security by making compressed models resilient to adversarial attacks such as bit-flip attacks. Sensitivity-aware compression increases the degradation of accuracy in tampered layers, facilitating effective tamper detection.
\end{itemize}

The remaining of this paper is structured as follows: Section II covers background and related work on model compression and weight generation, Section III details the WINGs framework and its dynamic weight generation process, Section IV presents experimental results, and Section V concludes with future research directions.


\vspace{-8pt}
\section{Background and Related Works}
\vspace{-4pt}
\subsection{Preliminaries}

\noindent\textbf{Principal component analysis (PCA):-} PCA \cite{pearson1901pca} reduces large datasets to fewer variables while retaining most of the information, improving processing speed, efficiency, and memory usage. 

\vspace{3pt}

\noindent \textbf{Support Vector Regression (SVR)} is a machine learning method used in various applications. 
SVR solves non-linear processes, which projects the original data into a high-dimensional space where it becomes linearly separable. The details of SVR are described in \cite{smola2004tutorial}.

\noindent\textbf{Bit-Flip Attack:-} Bit-Flip Attack (BFA) \cite{rahaman2024secure} is a potential adversarial attack that is launched on neural networks by maliciously flipping a small number of bits in weight memory. BFA can be easily launched on DNN weights stored in DRAM using the well-known row hammer attack (RHA), causing a significant drop in accuracy \cite{rakin2019bit} and can be difficult to detect.

\vspace{-4pt}
\subsection{Related Works}
Various methods have been explored to compress neural network weights to improve memory efficiency. 
Reducing weight bit-precision is an effective method to reduce both energy consumption and storage requirements \cite{Courbariaux_2015_ICLR}. However, compression is limited by minimum bit precision, with a 32-bit to 4-bit reduction offering only 8X compression. To achieve higher compression, strategies such as reducing the number of weights are used. Pruning is used to remove weights below a certain threshold \cite{hertz2018introduction} to reduce memory usage. The low-rank matrix approximation \cite{sainath2013low} reduces data representation by matrix factorization. Chen et al. applied run-length encoding for lossless compression of CNN output images \cite{chen2016eyeriss}. The conversion of the frequency domain \cite{koutnik2010evolving} is used to compress the weights by discarding the high-frequency components. Random hashing \cite{chen2016compressing} is used to group the frequency components of CNN filters into fewer hash buckets. 

Many compression methods require training modifications, such as fine-tuning to preserve accuracy. The method in \cite{han2015deep} involves fine-tuning (or retraining) after each compression stage to preserve the original accuracy. Techniques like binary weight training \cite{courbariaux2016binarized} can lead to accuracy drops if used without retraining. However, applying these techniques without retraining can cause accuracy drops \cite{chung2016simplifying}. An adaptive image encoding method \cite{ko2017adaptive} was introduced to compress weights without modifying the training process, focusing on the energy and performance of the system.

\vspace{-8pt}

\section{Methodology}
WINGs combines PCA and SVR to efficiently compress and dynamically reconstruct weights for DNNs. Fig.~\ref{fig:Lightweight} shows the process, which begins with a large weight matrix (Large Weight Table), which is reduced using PCA. This step identifies the most significant features by computing the covariance matrix and extracting eigenvectors associated with the largest eigenvalues, projecting the original high-dimensional weights into a compact subspace. The resulting compressed representation (Reduced Weight Table) retains the critical components while discarding less informative dimensions, thereby significantly reducing storage requirements. Once the dimensionality is reduced, SVRs are used to infer the weights for subsequent layers. These trained lightweight models (TLM1, TLM2, TLM3, etc.) sequentially generate the weights needed for each layer by using PCA-reduced components and the initial layer weights. This dynamic reconstruction eliminates the need to store the full weight matrix, enabling on-the-fly weight generation with minimal memory usage. During inference, the framework dynamically reconstructs the compressed weights for the low-sensitivity layers, while the high-sensitivity layers retain their original weights to ensure accuracy.

\begin{figure}[h]
    \centering
    \includegraphics[width=0.4\textwidth]{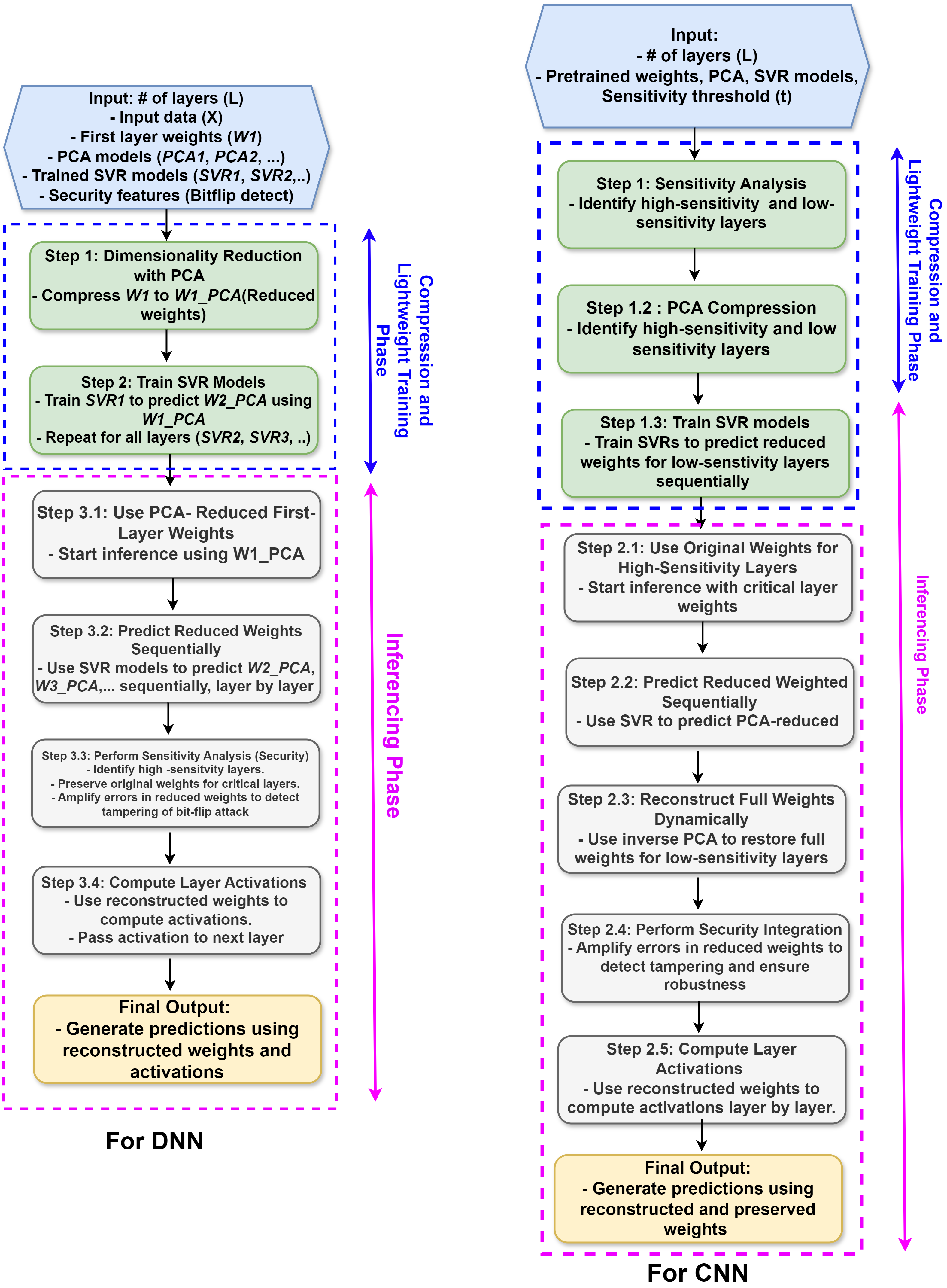}

    \caption {This figure shows a step-by-step overview of the WINGs weight generation workflow. It highlights the integration of principal component analysis (PCA) for dimensionality reduction and support vector regression (SVR) for dynamic reconstruction of layer weights. This enables efficient memory usage and inferencing of Deep Neural Networks.}
    \label{fig:Lightweight_1}
\end{figure}

The framework also incorporates a sensitivity-aware design, which uses gradient-based sensitivity analysis to determine which layers to compress. Low-sensitivity layers are selected for PCA-SVR compression, keeping high-sensitivity layers, which maintain model accuracy, remain unaffected. As depicted in Fig.~\ref{fig:Lightweight}, the sensitivity-aware structure allows the framework to target specific layers, minimizing accuracy loss even under resource constraints. Bit-flip attacks behave differently on compressed models than on original models. In the original model, parameter errors limit their propagation and ensure stability. However, as shown in Fig.~\ref{fig:Lightweight}, errors in the SVR predicted components propagate across layers due to the interdependence of parameters, increasing inaccuracies. This behavior, combined with the sensitivity-aware design, makes tampering more detectable. Fig.~\ref{fig:Impact of BitFlip Attacks on Original vs Compressed Model} illustrates how bit-flip attacks on compressed models lead to sharper accuracy degradation compared to original models, providing a built-in tamper detection mechanism. By combining PCA for dimensionality reduction, SVR for predictive reconstruction, and sensitivity-aware compression, the framework achieves memory efficiency, accuracy retention, and robustness against adversarial attacks. The step-by-step process for weight generation and compression for DNN Models is described in Figure \ref{fig:Lightweight_1}.


\vspace{-4pt}

\subsection{Weight Generation in Fully Connected Networks (FCNs)}

The proposed method operates in two phases: \textit{training} and \textit{inference}. In the \textit{training phase}, the weight matrix of each layer \(\mathbf{W}_\ell \in \mathbb{R}^{n_{\ell-1} \times n_\ell}\), where \(\mathbb{R}\) represents the set of real numbers, connects \(n_{\ell-1}\) neurons in layer \(\ell-1\) to \(n_\ell\) neurons in layer \(\ell\). PCA is applied to \(\mathbf{W}_\ell\), transforming it into a lower-dimensional representation \(\hat{\mathbf{W}}_\ell^{\text{PCA}} = \mathbf{U}_\ell^\top \mathbf{W}_\ell\), where \(\mathbf{U}_\ell \in \mathbb{R}^{n_{\ell-1} \times k_\ell}\) contains the top \(k_\ell\) eigenvectors of the covariance matrix of \(\mathbf{W}_\ell\), and \(k_\ell \ll n_{\ell-1}\) is chosen to retain at least 90\% of the variance. SVR models are trained to predict reduced weights of layer \(\ell+1\) from those of layer \(\ell\), minimizing the loss \(\mathcal{L}_\ell = \frac{1}{2} \|\hat{\mathbf{W}}_{\ell+1}^{\text{PCA}} - \text{SVR}_\ell(\hat{\mathbf{W}}_\ell^{\text{PCA}})\|_2^2 + C \sum_{i=1}^{k_\ell} \max(0, |\epsilon_i| - \epsilon)\), where \(C\) controls the trade-off between model complexity and error, and \(\epsilon\) is the error tolerance. This process is repeated for all layers, storing only the reduced weights of the first layer (\(\mathbf{W}_1\)) and trained SVR models.

In the \textit{inference phase}, stored data is used to dynamically reconstruct the weights for computation. The input \(\mathbf{X} \in \mathbb{R}^{m \times n_0}\), where \(m\) is the batch size and \(n_0\) is the number of input features, is used to compute the activations of the first layer as \(\mathbf{A}_1 = \sigma(\mathbf{X} \mathbf{W}_1)\), where \(\sigma(\cdot)\) is a non-linear activation function such as ReLU. For subsequent layers, PCA is applied to the weights of the previous layer to obtain their reduced form \(\hat{\mathbf{W}}_{\ell-1}^{\text{PCA}} = \text{PCA}_{\ell-1}(\mathbf{W}_{\ell-1})\), which are then used by the SVR model to predict the reduced weights for the current layer as \(\hat{\mathbf{W}}_\ell^{\text{PCA}} = \text{SVR}_{\ell-1}(\hat{\mathbf{W}}_{\ell-1}^{\text{PCA}})\). These predicted weights are reconstructed to their original dimensions via inverse PCA as \(\hat{\mathbf{W}}_\ell = \text{PCA}_\ell^{-1}(\hat{\mathbf{W}}_\ell^{\text{PCA}})\). The reconstructed weights compute the activations of the current layer as \(\mathbf{A}_\ell = \sigma(\mathbf{A}_{\ell-1} \hat{\mathbf{W}}_\ell)\). This process continues for all layers until the final output is computed using the softmax function as \(\mathbf{A}_L = \text{Softmax}(\mathbf{A}_{L-1} \hat{\mathbf{W}}_L)\). The PCA-SVR combination generates on-the-fly weights and avoids storing full weight matrices while maintaining high accuracy. Algorithm 1 depicts the training and interference phase.

\begin{algorithm}[ht]
    \caption{Layer-wise Weight Prediction with PCA and Lightweight Models}
    \label{alg:weight_prediction}
    \begin{algorithmic}[1]
        \State \textbf{Input:} $L$: Number of layers, $\mathbf{X} \in \mathbb{R}^{m \times n_0}$: Input data, $\mathbf{W}_1$: Stored first-layer weights, $\{\text{SVR}_\ell\}_{\ell=2}^{L}$: Trained SVR models, $\{\text{PCA}_\ell\}_{\ell=1}^{L-1}$: PCA models
        \State \textbf{Output:} $\mathbf{A}_L$: Final layer activations
        \vspace{0.2em}
        \State \textbf{Training Phase:}
        \For {$\ell = 1$ to $L-1$}
            \State Apply PCA to reduce $\mathbf{W}_\ell$: $\hat{\mathbf{W}}_\ell^{\text{PCA}} = \text{PCA}_\ell(\mathbf{W}_\ell)$
            \State Train $\text{SVR}_\ell$ to map reduced weights: $\hat{\mathbf{W}}_\ell^{\text{PCA}} \mapsto \hat{\mathbf{W}}_{\ell+1}^{\text{PCA}}$
        \EndFor
        \State Store $\mathbf{W}_1$ and discard all other $\mathbf{W}_\ell$
        \vspace{0.2em}
        \State \textbf{Inference Phase:}
        \State Compute first-layer activations: $\mathbf{A}_1 = \sigma(\mathbf{X} \mathbf{W}_1)$
        \For {$\ell = 2$ to $L$}
            \State Predict reduced weights for layer $\ell$: 
            \[
            \hat{\mathbf{W}}_\ell^{\text{PCA}} = \text{SVR}_{\ell-1}(\hat{\mathbf{W}}_{\ell-1}^{\text{PCA}})
            \]
            \State Reconstruct full weights: $\hat{\mathbf{W}}_\ell = \text{PCA}_\ell^{-1}(\hat{\mathbf{W}}_\ell^{\text{PCA}})$
            \State Compute activations: $\mathbf{A}_\ell = \sigma(\mathbf{A}_{\ell-1} \hat{\mathbf{W}}_\ell)$
        \EndFor
        \State Compute final output: $\mathbf{A}_L = \text{Softmax}(\mathbf{A}_{L-1} \hat{\mathbf{W}}_L)$
    \end{algorithmic}
\end{algorithm}

\vspace{-4pt}
\subsection{Weight Compression and Sensitivity Analysis in CNNs}
The proposed method compresses CNNs by identifying low-sensitivity layers for targeted compression. To minimize accuracy loss, the process begins with sensitivity analysis, where the impact of weight changes on the loss function is quantified. For each layer \(W_l\), the sensitivity is calculated as \(S_l = \mathbb{E}_{(x, y) \sim \mathcal{D}} \left[ \| \nabla_{W_l} \mathcal{L}(x, y) \|_F \right]\), where \(\mathcal{D}\) is the dataset, \((x, y)\) are input-output pairs, \(\mathcal{L}\) is the loss function, and \(\|\cdot\|_F\) is the Frobenius norm \cite{golub2013matrix}. This calculation measures how changes in the weights of each layer affect the performance of the model, with higher gradient norms indicating higher sensitivity. Layers with sensitivity \(S_l < \tau\) are identified as low-sensitivity layers, which are less critical for maintaining model accuracy and therefore suitable for compression. The identified layers are grouped into a set \(\mathcal{L}_{\text{selected}} = \{W_l : S_l < \tau\}\), allowing the framework to focus computational efforts on compressing layers that tolerate changes without significant accuracy degradation.

Once low-sensitivity layers are identified, the framework employs Principal Component Analysis (PCA) to reduce the dimensionality of their weight matrices. PCA works by transforming each weight matrix \(W_l\) into a lower-dimensional space while retaining the most significant information. To begin with, the weight matrix is centered by subtracting its mean vector \(\mu_l\), resulting in \(\tilde{W}_l = W_l - \mathbf{1}_m \mu_l^\top\), where \(\mathbf{1}_m\) is a vector of length \(m\). The covariance matrix \(\Sigma_l = \frac{1}{m} \tilde{W}_l^\top \tilde{W}_l\) is then computed to capture the variance and relationships between the weights. The decomposition of the eigenvalues of \(\Sigma_l\) provides the eigenvalues \(\lambda_i\) and the eigenvectors, and the upper \(k\) eigenvectors, stored in \(P_l\), are selected to satisfy the variance retention ratio \(\eta = \frac{\sum_{i=1}^k \lambda_i}{\sum_{j=1}^n \lambda_j}\). This ensures that the compressed representation, \(W_l^{\text{PCA}} = \tilde{W}_l P_l\), retains the most important features while reducing the number of parameters.

\begin{algorithm}[ht]
\caption{Gradient-Based Sensitivity-Aware PCA + SVR Compression}
\label{alg:sensitivity_aware}
\begin{algorithmic}[1]
\Require Model \( \mathcal{M} \) with layers \( \{W_1, W_2, \dots, W_L\} \), sensitivity threshold \( \tau \), PCA ratio \( \rho \), SVR configuration
\Ensure Compressed weights \( \{\hat{W}_i\} \)

\State \textbf{1. Input:} Trained model \(\mathcal{M}\), \(\tau\), PCA ratio \(\rho\), SVR configuration
\State \textbf{2. Output:} Compressed weights \(\{\hat{W}_i\}\)
\vspace{0.2em}

\State \textbf{Step 1: Sensitivity Computation:} Compute \( S_i = \mathbb{E}_{(x, y)} \| \nabla_{W_i} \mathcal{L}(x, y) \|_F \quad \forall W_i \).

\State \textbf{Step 2: Layer Selection:} Select \(\mathcal{L}_{\text{selected}} = \{ W_i : S_i < \tau \}\).

\For{each \( W_i \in \mathcal{L}_{\text{selected}} \)}
    \State \textbf{Step 3: PCA Compression:}
    \begin{itemize}
        \item Center: \( \tilde{W}_i = W_{\text{flat}}^{(i)} - \mathbf{1} \mu_i^T \)
        \item Covariance: \( \Sigma_i = \frac{1}{C_{\text{out}}} \tilde{W}_i^T \tilde{W}_i \)
        \item Top eigenvectors: \( P_i^k \), compressed weights: \( W_i^{\text{PCA}} = \tilde{W}_i P_i^k \)
    \end{itemize}

    \State \textbf{Step 4: SVR Prediction:}
    \begin{itemize}
        \item Split: \( W_i^{\text{PCA}} = [W_i^{\text{known}}, W_i^{\text{predict}}] \)
        \item Predict: \( \hat{W}_i^{\text{predict}}[:, j] = f_{i,j}(W_i^{\text{known}}) \)
    \end{itemize}

    \State \textbf{Step 5: Reconstruction:}
    \begin{itemize}
        \item Combine: \( \hat{W}_i^{\text{PCA}} = \begin{bmatrix} W_i^{\text{known}} & \hat{W}_i^{\text{predict}} \end{bmatrix} \)
        \item Inverse PCA: \( \hat{W}_{\text{flat}}^{(i)} = \hat{W}_i^{\text{PCA}} (P_i^k)^T + \mathbf{1} \mu_i^T \)
        \item Reshape: \( \hat{W}_i \)
    \end{itemize}
\EndFor

\State \textbf{Step 6: Inference:} Use \(\hat{W}_i\) for low-sensitivity layers (\(W_i \in \mathcal{L}_{\text{selected}}\)), and \(W_i\) otherwise.

\State \textbf{Step 7: Attack Awareness:} Errors in \(\hat{W}_i\) (compressed layers) propagate but are amplified, making them easier to detect.
\end{algorithmic}
\end{algorithm}

However, dimensionality reduction via PCA discards less significant components, potentially leading to information loss. To address this, SVR is applied to predict the omitted parts of the compressed weights. Each reduced weight matrix \(W_l^{\text{PCA}}\) is divided into a known part \(W_l^{\text{known}}\) and a missing part \(W_l^{\text{predict}}\). For each column of \(W_l^{\text{predict}}\), an SVR model is trained in \(W_l^{\text{known}}\) to predict the missing values, such that \(\hat{W}_l^{\text{predict}}[:, j] = \text{SVR}_j(W_l^{\text{known}})\). This predictive modeling ensures that even omitted components are accurately reconstructed, mitigating the effects of compression on model performance.

During inference, the framework dynamically reconstructs the weight matrices to their full dimensions. The reconstructed weights are obtained by combining the known and predicted components using inverse PCA: \(\hat{W}_l = W_l^{\text{compressed}} P_l^\top + \mu_l\), where \(W_l^{\text{compressed}}\) integrates both the predicted and retained components. This reconstruction process ensures that the compressed weights retain their essential characteristics while supporting efficient forward propagation through the network.

Despite the advantages of compression, reduced redundancy in the weight matrices introduces potential vulnerabilities to bit-flip attacks. In the compressed model, errors in weights that occur can propagate across layers due to the interconnected nature of SVR-predicted components, increasing their impact on accuracy and enhancing the detectability of attacks. By combining PCA for dimensionality reduction and SVR for predictive reconstruction, the framework achieves significant memory savings while maintaining accuracy and robustness for practical deployment.

\vspace{-4pt}
\section{Security Analysis}

WINGs significantly improves the security of DNNs by increasing attack complexity, enhancing detectability, and reducing protection costs. 


\vspace{-8pt}

\subsection{Increased Attack Complexity}
Weights in WINGs are reconstructed as $W \approx Z \cdot P^T + \mu$, where $Z$ is the PCA-reduced representation, $P$ is the projection matrix, and $\mu$ is the mean vector. Attackers must reverse-engineer both the PCA compression and SVR prediction processes. The attack complexity grows as:
\[
C_{\text{attack}} = \prod_{i=1}^{n_{\text{layers}}} \left(C_{\text{PCA},i} \cdot C_{\text{SVR},i}\right),
\]
where $C_{\text{PCA},i}$ and $C_{\text{SVR},i}$ represent the efforts required to obtain PCA and SVR parameters for layer $i$. Cascading dependencies among layers further amplify the difficulty of reconstructing the weights, making WINGs highly resilient to such attacks.


\vspace{-4pt}

\subsection{Enhanced Detectability}
In WINGs, a bit-flip $\delta$ in the predicted component $Z_{\text{pred}}$ propagates through reconstruction as:
\[
W' = ((Z_{\text{pred}} + \delta) \oplus Z_{\text{known}}) \cdot P^T + \mu.
\]
It amplifies accuracy degradation, with the error amplification factor:
\[
A = \frac{\Delta \text{Accuracy}_{\text{compressed}}}{\Delta \text{Accuracy}_{\text{original}}}, \quad A \gg 1.
\]
Compressed models exhibit sharper accuracy drops, as shown in \ref{fig:Impact of BitFlip Attacks on Original vs Compressed Model}.


\vspace{-8pt}

\subsection{Reduced Protection Costs}
Protecting weights against bit-flip attacks using ECC incurs an 18\% overhead per bit. WINGs reduces this cost by compressing weights by up to 92\%. For a model with $N_{\text{original}}$ weights, the ECC cost is below (sensitivity-aware protection further lowers costs).
\[
\text{Protection Cost}_{\text{compressed}} = \text{Protection Cost}_{\text{original}} \times (1 - 0.92).
\]


\vspace{-4pt}
\section{Experimental Results and Discussions}
The proposed method is evaluated in multiple data sets to assess its effectiveness in reducing memory usage with minimum accuracy loss. In this section, we present the detailed results, which apply specifically to fully connected layers and convolutional layers, and discuss the sensitivity-based attacks such as bit-flip attacks.

\begin{table*}[t]
\caption{Summary of model sizes, accuracy, and PCA impact across datasets}
\label{tab:my-table2}
\footnotesize 
\resizebox{\textwidth}{!}{%
\begin{tabular}{|l|c|c|c|c|c|c|c|c|c|c|c|}
\hline
\textbf{Dataset} &
  \textbf{\begin{tabular}[c]{@{}c@{}}Input \\ Size\end{tabular}} &
  \textbf{\begin{tabular}[c]{@{}c@{}}Orig. \\ Model \\ (KB)\end{tabular}} &
  \textbf{\begin{tabular}[c]{@{}c@{}}1st Layer \\ Weight \\ (KB)\end{tabular}} &
  \textbf{\begin{tabular}[c]{@{}c@{}}PCA-Red. \\ 1st Layer \\ (KB)\end{tabular}} &
  \textbf{\begin{tabular}[c]{@{}c@{}}Light \\ Model \\ (KB)\end{tabular}} &
  \textbf{\begin{tabular}[c]{@{}c@{}}Mod. \\ Model \\ (KB)\end{tabular}} &
  \textbf{\begin{tabular}[c]{@{}c@{}}Size \\ Red. \\ (KB)\end{tabular}} &
  \textbf{\begin{tabular}[c]{@{}c@{}}Size \\ Red. \\ (\%)\end{tabular}} &
  \textbf{\begin{tabular}[c]{@{}c@{}}Orig. \\ Acc. \\ (\%)\end{tabular}} &
  \textbf{\begin{tabular}[c]{@{}c@{}}Light \\ Acc. \\ (\%)\end{tabular}} &
  \textbf{\begin{tabular}[c]{@{}c@{}}Acc. \\ Red. \\ (\%)\end{tabular}} \\ \hline
\textbf{SVHN} &
  32x32 &
  6822.29 &
  6144 &
  920 &
  131.08 &
  1051.08 &
  5771.21 &
  84.59 &
  79.13 &
  77.20 &
  1.93 \\ \hline
\textbf{CNAE-9} &
  856D &
  2390.04 &
  1712 &
  920 &
  121.96 &
  1041.96 &
  1348.07 &
  56.40 &
  90.74 &
  87.04 &
  3.70 \\ \hline
\textbf{Fashion-MNIST} &
  28x28 &
  2246.29 &
  1568 &
  921.80 &
  123.27 &
  1045.07 &
  1201.22 &
  53.48 &
  87.85 &
  87.38 &
  0.47 \\ \hline
\textbf{MNIST} &
  28x28 &
  2246.29 &
  1568 &
  920 &
  128.53 &
  1048.53 &
  1197.76 &
  53.32 &
  97.56 &
  97.22 &
  0.34 \\ \hline
\end{tabular}%
}
\end{table*}

\noindent\textbf{Results on model sizes, accuracy and PCA impact on datasets:-}
Table \ref{tab:my-table2} presents the original model size, PCA-reduced first layer size, compressed size, etc., using four different data sets:SVHN, CNAE-9, Fashion MNIST, and MNIST. The results show that SVHN achieves a reduction in model size of 84.59\%, while CNAE-9, Fashion MNIST, and MNIST demonstrate reductions of 56.40\%, 53.48\% and 53.32\%, respectively, but the CNAE-9 dataset shows the largest decrease in accuracy of 3.70\%, while the MNIST and Fashion MNIST data sets retain their original accuracy of 97\% with decreases in accuracy of only 0.34\% and 0. 47\%, respectively. This demonstrates the effectiveness of PCA-SVR compression in maintaining accuracy. It retains high-sensitivity layers uncompressed to preserve critical information and minimize accuracy loss.

\noindent\textbf{Impact of PCA thresholds on model accuracy and compression:}

\begin{figure}[!h]
    \centering
    \includegraphics[width=0.8\columnwidth]{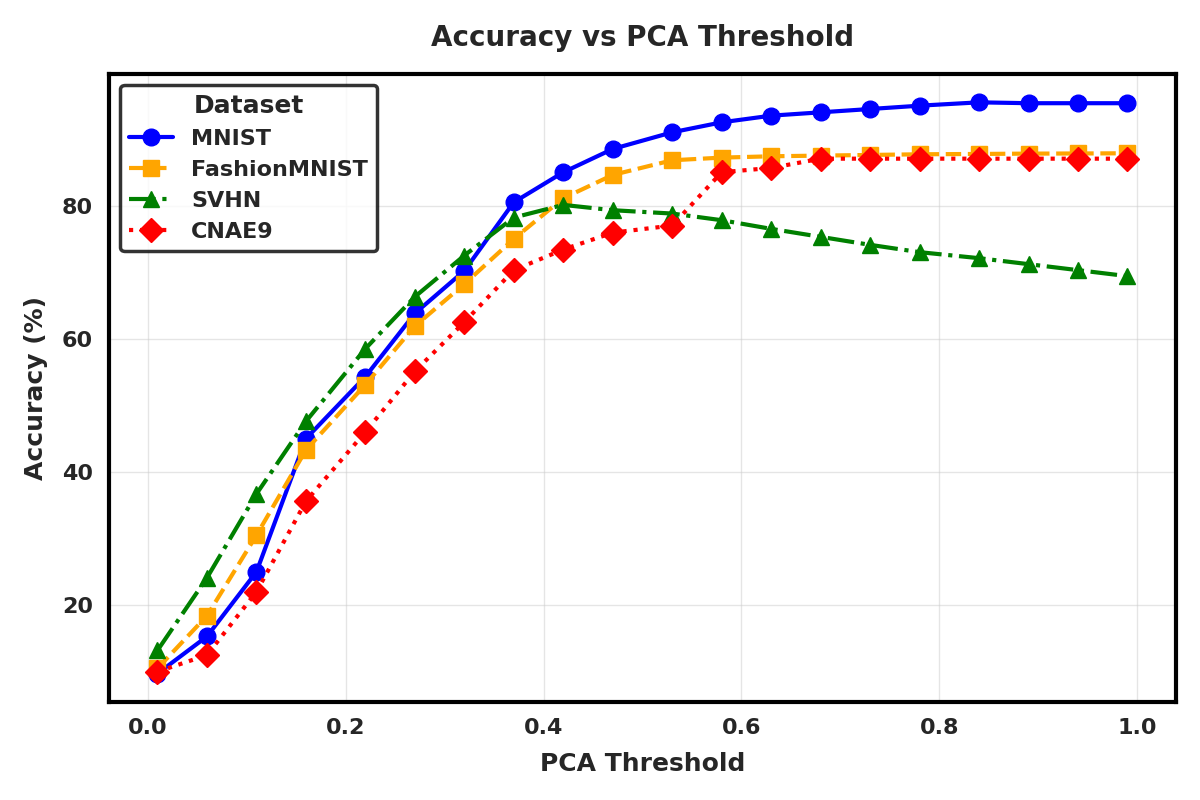}

    \caption{This figure shows the impact of PCA thresholds on model accuracy. Here we have analyzed how varying dimensionality reduction levels affect accuracy retention across different datasets. This highlights the trade-off between compression and performance.}
    \label{fig:AP_Threshold}
\end{figure}

\begin{figure}[!h]
    \centering
    \includegraphics[width=0.8\columnwidth]{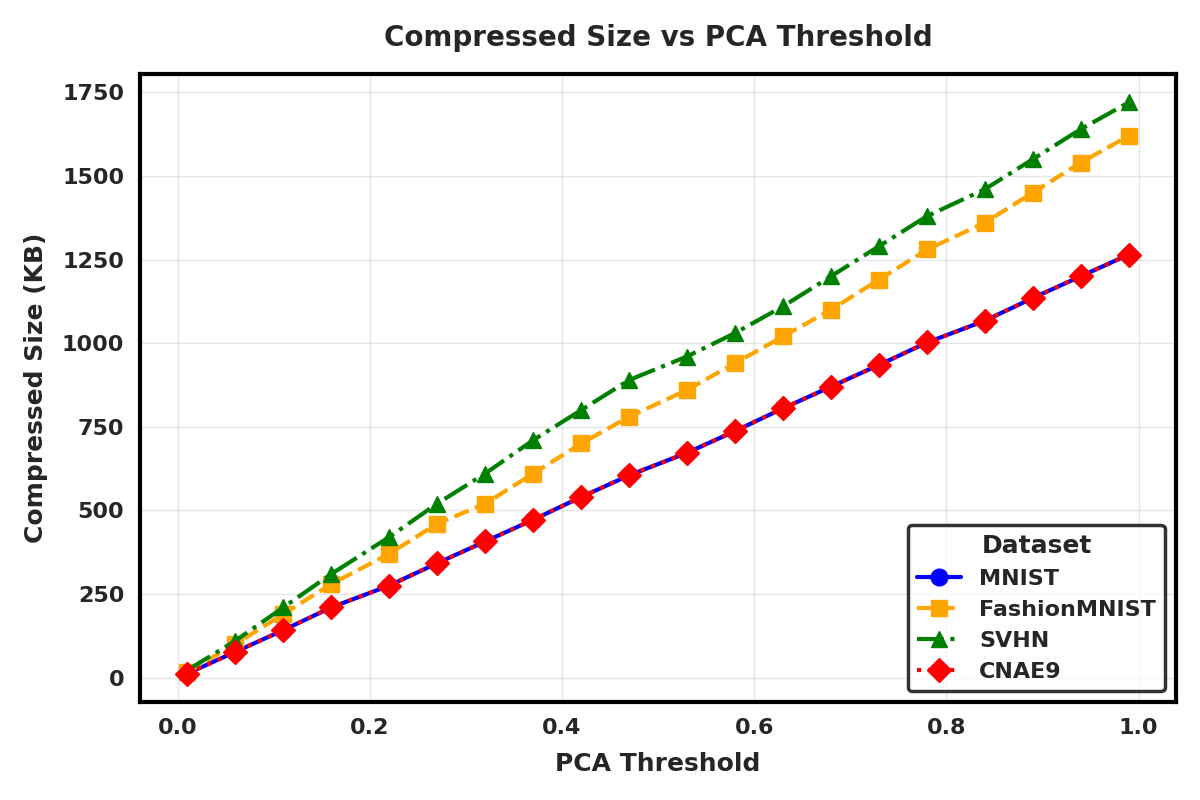}
    
    \caption{This figure shows the effect of PCA thresholds on compressed model size. It illustrates the relationship between PCA thresholds and memory savings, demonstrating significant size reductions achievable with lower thresholds.}
    \label{fig:Compressed_Threshold}
\end{figure}

Figures \ref{fig:AP_Threshold}, and Figure \ref{fig:Compressed_Threshold} explore the impact of PCA thresholds on accuracy and compression in various data sets. Lower thresholds yield higher compression with reduced accuracy, while higher thresholds retain more accuracy with less compression. It shows that at a PCA threshold of 0.06, MNIST achieves a size reduction of 96. 57\% but only with a precision of 15.3\%, while at 0.99, it retains an accuracy of 95.37\% with a size reduction of 43.64\%. Data sets exhibit varying sensitivity to PCA thresholds. MNIST and Fashion MNIST are relatively resistant to compression for accuracy loss, while SVHN and CNAE-9 require higher thresholds for stable accuracy.


\begin{figure}[!h]
    \centering
    \includegraphics[width=0.8\columnwidth]{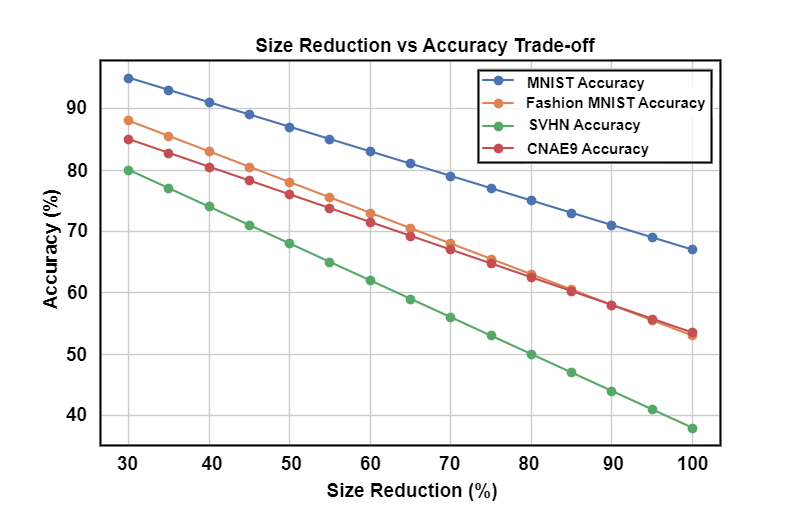}
    \caption{This figure shows the trade-off between size reduction and accuracy retention. It analyzes dataset-specific compression limits with PCA-SVR framework.}
    \label{fig:Reduction_Accuracy}
\end{figure}

Figure \ref{fig:AP_Threshold} highlights the optimal PCA thresholds specific to the data set, with thresholds (0.5-0.7) that achieve precision above 80\% and significant compression for MNIST and Fashion MNIST. Figure \ref{fig:Compressed_Threshold} illustrates that lower PCA thresholds (0.1–0.4) provide significant size reductions, suitable for memory-constrained applications. Compressed model sizes for MNIST, Fashion MNIST, SVHN, and CNAE-9 drop substantially at these thresholds, while higher thresholds offer a near-linear size increase, enabling fine-tuned control for specific memory and application requirements. Figure \ref{fig:Reduction_Accuracy} highlights the impact of size reduction and accuracy trade-off. MNIST and Fashion MNIST show greater resistance to accuracy loss, while SVHN and CNAE-9 need less compression to preserve performance, highlighting dataset-specific compression limits.

\vspace{-5pt}

\vspace{-3pt}

\vspace{-8pt}
\subsection{Overall Compression and Accuracy}
The algorithm 2 is performed on Alexnet with the CIFAR10 [Table~\ref{tab:alexnet_cifar10}] and MINST [Table \ref{tab:alexnet_mnist}] datasets for the fully connected layers (FC) of AlexNet. Both the tables [\ref{tab:alexnet_cifar10} and \ref{tab:alexnet_mnist}] show aggressive compression using PCA and SVR, achieving substantial reductions in memory usage while maintaining accuracy. Table~\ref{tab:alexnet_cifar10}, shows that FC6 and FC7 layers achieved a compression ratio of 28.80$\times$. FC8 layer remained uncompressed due to its small size and high sensitivity. Using sensitivity-aware compression, the WINGs selectively targeted low-sensitivity convolutional (conv) layers (conv4 and conv5) for compression, while retaining the original weights for high-sensitivity layers. Table~\ref{tab:alexnet_cifar10} presents the compression for AlexNet's conv layers. The results show significant compression for conv4 and conv5, with a compression ratio of 28.8$\times$. These layers were compressed with minimal impact on feature extraction. It reduced AlexNet's total size from 136.75 MB to 7.63 MB, with a compression ratio of 17.92$\times$.

\begin{table}[ht]
\centering
\resizebox{0.8\columnwidth}{!}{ 
\begin{tabular}{|l|c|c|c|}
\hline
\textbf{Layer} & \textbf{Original Size (MB)} & \textbf{Compr. Size (MB)} & \textbf{Compr. Ratio} \\ \hline
Conv1 & 0.0061 & 0.000244 & 25.0164 \\ \hline
Conv2 & 0.4220 & 0.013916 & 30.3248 \\ \hline
Conv3 & 2.5310 & 0.087891 & 28.797 \\ \hline
Conv4 & 3.3750 & 0.117188 & 28.7999 \\ \hline
Conv5 & 2.2500 & 0.078125 & 28.8 \\ \hline
FC6   & 36.0000 & 1.2500   & 28.8 \\ \hline
FC7   & 64.0000 & 2.21875  & 28.8451 \\ \hline
FC8   & 0.1560  & 0.0267   & 5.85206 \\ \hline
\textbf{Total} & \textbf{108.7404} & \textbf{3.7928} & \textbf{28.6701} \\ \hline
\end{tabular}
}
\caption{Comparison results for AlexNet on MNIST dataset denoting original and compressed size for convolutional and fully connected layers, along with the achieved compression ratios using the WINGs framework}
\label{tab:alexnet_mnist}
\end{table}


\begin{table}[ht]
\centering
\resizebox{0.8\columnwidth}{!}{ 
\begin{tabular}{|l|c|c|c|}
\hline
\textbf{Layer} & \textbf{Original Size (MB)} & \textbf{Compr. Size (MB)} & \textbf{Compr. Ratio} \\ \hline
Conv1 & 0.0183 & 0.018   & 1       \\ \hline
Conv2 & 0.4220 & 0.422   & 1       \\ \hline
Conv3 & 2.5300 & 2.530   & 1       \\ \hline
Conv4 & 3.3750 & 0.117   & 28.797  \\ \hline
Conv5 & 2.2500 & 0.078   & 28.800  \\ \hline
FC6   & 64.0000 & 2.219   & 28.842  \\ \hline
FC7   & 64.0000 & 2.219   & 28.842  \\ \hline
FC8   & 0.1560  & 0.0267  & 5.852   \\ \hline
\textbf{Total} & \textbf{136.752} & \textbf{7.630} & \textbf{17.922} \\ \hline
\end{tabular}
}
\caption{AlexNet with CIFAR-10 Compression Results}
\label{tab:alexnet_cifar10}
\end{table}

The proposed PCA-SVR-based method is evaluated using three datasets: MNIST, CNAE-9, and SVHN in a fully connected network with three and four layers. Performance comparisons in Table 1 include other compression techniques such as RLE and Huffman lossless\cite{han2015deep, chen2016eyeriss}, pruning \cite{han2015deep, hertz2018introduction}, truncation \cite{courbariaux2015low}, and JPEG compression \cite{ko2017adaptive} using 32-bit fixed point weights. 

\vspace{-3 pt}

\begin{table}[!h]
\centering
\caption{Compression ratios of various techniques applied to different datasets and comparing the performance of traditional methods with the proposed WINGs framework}
\label{tab:my-table}
\resizebox{0.8\columnwidth}{!}{
\begin{tabular}{|c|c|c|c|c|c|c|c|c|}
\hline
\textbf{Dataset} &
  \textbf{\begin{tabular}[c]{@{}c@{}}Baseline\\ Accuracy\end{tabular}} &
  \textbf{Layer} &
  \textbf{\begin{tabular}[c]{@{}c@{}}Size\\ (KB)\end{tabular}} &
  \textbf{\begin{tabular}[c]{@{}c@{}}RLE+ \\ Huffman \\  \cite{han2015deep}\cite{chen2016eyeriss}\end{tabular}} &
  \textbf{\begin{tabular}[c]{@{}c@{}}Pruning \\ \cite{han2015deep, hertz2018introduction} \end{tabular}} &
  \textbf{\begin{tabular}[c]{@{}c@{}}Trunc-\\ ation \\ \cite{courbariaux2015low} \end{tabular}} &
  \textbf{\begin{tabular}[c]{@{}c@{}}JPEG \\ Compression\cite{ko2017adaptive}\end{tabular}} &
  \textbf{Proposed} \\ \hline
\multirow{3}{*}{MNIST}  & \multirow{3}{*}{97.80\%} & FC1   & 441   & 1.1 & 5.9 & 8 & 64.3 & 52.25 \\ \cline{3-9} 
                        &                          & FC2   & 5.6   & 1.1 & 2.9 & 8 & 30.1 & 71.80 \\ \cline{3-9} 
                        &                          & Total & 446.6 & 1.1 & 5.9 & 8 & 63.4 & 52.48 \\ \hline
\multirow{4}{*}{CNAE-9} & \multirow{4}{*}{97.20\%} & FC1   & 481.5 & 1.1 & 4.5 & 8 & 26.5 & 24.67 \\ \cline{3-9} 
                        &                          & FC2   & 36    & 1.1 & 3.2 & 8 & 15.5 & 36.27 \\ \cline{3-9} 
                        &                          & FC3   & 2.3   & 1.0 & 2.1 & 8 & 5.50 & 41.07 \\ \cline{3-9} 
                        &                          & Total & 519.8 & 1.1 & 4.4 & 8 & 24.8 & 26.19 \\ \hline
\multirow{4}{*}{SVHN}   & \multirow{4}{*}{76.90\%} & FC1   & 576.0 & 1.2 & 3.1 & 4 & 42.3 & 40.56 \\ \cline{3-9} 
                        &                          & FC2   & 36.0  & 1.1 & 1.6 & 4 & 6.80 & 28.80 \\ \cline{3-9} 
                        &                          & FC3   & 2.5   & 1.1 & 1.9 & 4 & 4.90 & 32.07 \\ \cline{3-9} 
                        &                          & Total & 614.5 & 1.2 & 3.0 & 4 & 31.6 & 39.57 \\ \hline
\end{tabular}%
}
\end{table}

\vspace{-5pt}
\subsection{Sensitivity-Aware Compression and Bit-Flip Attack Resilience}

\begin{figure}[h]
    \centering
    \includegraphics[width=0.9\columnwidth]{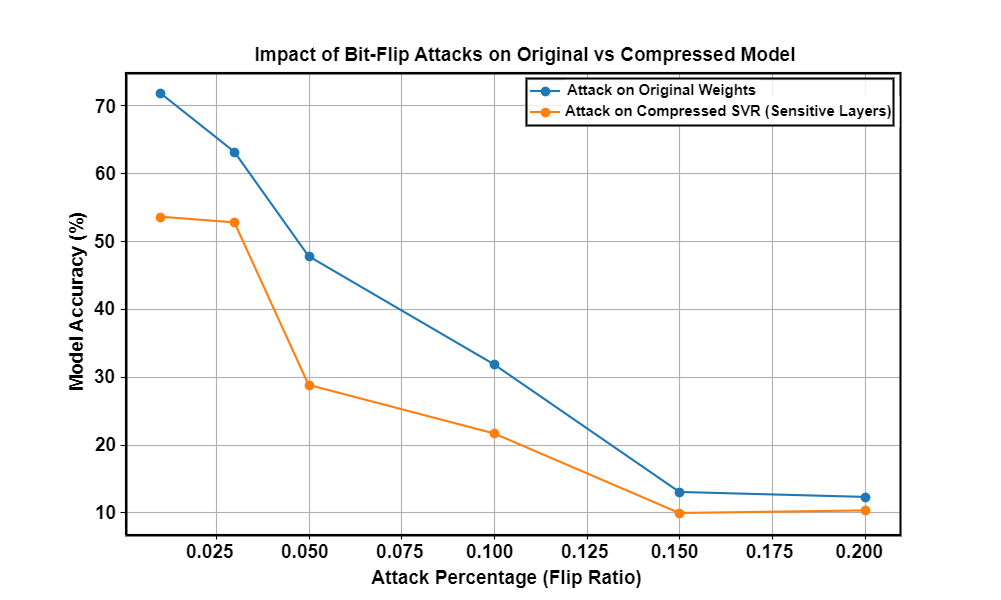}

    \caption{This figure shows the impact of bit-flip attacks on model accuracy. It compares the robustness of the original and the PCA-SVR compressed models. It can be observed that there is a sharper accuracy degradation in compressed models under adversarial tampering.}
    \label{fig:Impact of BitFlip Attacks on Original vs Compressed Model}
\end{figure}

Under bit-flip attacks on the internal parameters of the SVR, the compressed model exhibits a post-bit-flip accuracy of 62\%, compared to 67.27\% for general compressed weights. This significant accuracy drop highlights the detectability of such attacks, ensuring tamper resistance. Figure~\ref{fig:Impact of BitFlip Attacks on Original vs Compressed Model} compares the accuracy between the original and compressed models with increasing levels of bit-flip attack. 


Figure~\ref{fig:Impact of BitFlip Attacks on Original vs Compressed Model} illustrates that compressed models experience steeper accuracy drops than original models as the intensity of the bit-flip attack increases. As compressed models are more vulnerable to such attacks, this sensitivity also enhances tamper detection. Even minor errors in compressed layers significantly degrade accuracy, enabling built-in tamper detection.

\vspace{-10pt}

\section{Conclusion}
We have presented \textbf{WINGs}, a framework for compressing and dynamically generating layer weights during inference, eliminating the need to store full-weight matrices, resulting in a significant reduction in storage needs without sacrificing accuracy. \textit{WINGs} leverages \textit{PCA} for dimensionality reduction and \textit{(SVR)} to predict and reconstruct compressed weights efficiently. For the first time, WINGS simultaneously addresses model storage efficiency for resource-constrained applications and attack-resistant (with respect to bit-flip attacks) inferencing at minimal impact on accuracy. The \textit{sensitivity-aware preferential compression} technique in \textbf{WINGs} uses gradient-based sensitivity analysis to selectively compress low-sensitivity layers, while preserving high-sensitivity layers in their original form to maintain model accuracy. The proposed approach integrates innovative weight generation, model compression, and sensitivity-aware compression to enhance robustness against adversarial attacks such as bit-flip attacks. We show significant reduction in memory requirement, which results in higher throughput and lower energy for DNN inference, making it ideal for real-time applications, such as edge-based intelligent computing. Future directions include expanding WINGs to tackle more complex tasks, such as multimodal and unsupervised learning, where efficiency and scalability are essential.

\bibliographystyle{IEEEtranN}
\bibliography{bibliography}

\end{document}